\documentclass{article}

\usepackage{PRIMEarxiv}

\usepackage[utf8]{inputenc} 
\usepackage[T1]{fontenc}    
\usepackage{hyperref}       
\usepackage{url}            
\usepackage{booktabs}       
\usepackage{amsfonts}       
\usepackage{nicefrac}       
\usepackage{microtype}      
\usepackage{lipsum}
\usepackage{fancyhdr}       
\usepackage{graphicx}       
\usepackage{amsmath,amssymb,amsfonts}
\usepackage{cite, algorithmic, textcomp}
\usepackage{arydshln}
\usepackage{multirow}

\usepackage{amsthm}%
\usepackage{mathrsfs}%
\usepackage[title]{appendix}%
\usepackage{xcolor}%
\usepackage{tikz}
\usetikzlibrary{automata, positioning}
\usepackage{orcidlink}
\usepackage{arydshln}

\usepackage{subcaption}  
\usepackage{cleveref}
\usepackage{soul}
\usepackage{natbib}

\title{Artificial Inductive Bias for Synthetic Tabular Data Generation in Data-Scarce Scenarios}

\author{
  Patricia A. Apellániz\thanks{\textbf{These authors contributed equally.}}\\
  Information Processing \\and Telecommunications Center \\
  ETS Ingenieros de Telecomunicación \\
  Universidad Politécnica de Madrid \\
  Madrid\\
  \textit{patricia.alonsod@upm.es} \\
  \And
  Ana Jim\'enez{\textsc {$^*$}}\\
  Information Processing \\and Telecommunications Center \\
  ETS Ingenieros de Telecomunicación \\
  Universidad Politécnica de Madrid \\
  Madrid\\
  \textit{ana.jimenezb@upm.es} \\
  \And
  Borja Arroyo Galende\\
  Information Processing \\and Telecommunications Center \\
  ETS Ingenieros de Telecomunicación \\
  Universidad Politécnica de Madrid \\
  Madrid\\
  \textit{borja.arroyog@upm.es} \\
  \And
  Juan Parras\\
  Information Processing \\and Telecommunications Center \\
  ETS Ingenieros de Telecomunicación \\
  Universidad Politécnica de Madrid \\
  Madrid\\
  \textit{j.parras@upm.es} \\
  \And
  Santiago Zazo\\
  Information Processing \\and Telecommunications Center \\
  ETS Ingenieros de Telecomunicación \\
  Universidad Politécnica de Madrid \\
  Madrid\\
  \textit{santiago.zazo@upm.es} \\
}

\begin{document}

\maketitle

\begin{abstract}
While synthetic tabular data generation using Deep Generative Models (DGMs) offers a compelling solution to data scarcity and privacy concerns, their effectiveness relies on the availability of substantial training data, often lacking in real-world scenarios. To overcome this limitation, we propose a novel methodology that explicitly integrates artificial inductive biases into the generative process to improve data quality in low-data regimes. Our framework leverages transfer learning and meta-learning techniques to construct and inject informative inductive biases into DGMs. We evaluate four approaches (pre-training, model averaging, Model-Agnostic Meta-Learning (MAML), and Domain Randomized Search (DRS)) and analyze their impact on the quality of the generated text. Experimental results show that incorporating inductive bias substantially improves performance, with transfer learning methods outperforming meta-learning, achieving up to 60\% gains in Jensen-Shannon divergence. The methodology is model-agnostic and especially relevant in domains such as healthcare and finance, where high-quality synthetic data are essential, and data availability is often limited.
\end{abstract}


\section{Introduction}
\label{intro}
Recent advances in Deep Learning (DL) have led to the development of powerful Deep Generative Models (DGMs). These models excel at learning and representing complex, high-dimensional distributions, allowing them to sample new data points realistically. This capability has driven remarkable progress in various domains, including image generation \citep{elasri2022image}, text generation \citep{zhang2023survey}, and video generation \citep{selva2023video}. 

Tabular data has gained increasing interest within DGMs due to their structured format, making them fundamental for information storage and analysis across various fields. Researchers are actively investigating techniques for generating realistic and informative tabular data, as evidenced by the growing work in this area \citep{borisov2022deep, fonseca2023tabular}. Since the introduction of Generative Adversarial Networks (GANs) \citep{Goodfellow2014GenerativeAN}, a wide range of GAN-based models have been proposed for generating synthetic tabular data. Among them, CTGAN \citep{xu2019modeling} is one of the most prominent and widely adopted models, as it introduces conditional generation mechanisms that allow it to handle the multimodal, imbalanced, and mixed-type nature of tabular data. Building on CTGAN, CTAB-GAN \citep{zhao2021ctab} further improves the generation of both categorical and continuous variables with complex distributions. GANBLR \citep{zhang2021ganblr} proposes an interpretable GAN-based generator tailored to categorical data, achieving strong performance in both statistical similarity and downstream utility. Another notable architecture is TableGAN \citep{park2018data}, which enhances vanilla GANs by incorporating classification and information loss objectives to improve the semantic and statistical alignment between real and synthetic samples. However, TableGAN is less effective at handling categorical imbalances and lacks the conditional generation capabilities of CTGAN-based models. In addition, the ACGAN \citep{odena2017conditional} extends the discriminator to include an auxiliary classification task, enabling class-aware generation and improving semantic consistency. While originally designed for in-distribution generation, ACGAN has served as the foundation for more recent frameworks that generalize beyond the training distribution. For instance, \citep{jian2024open} proposes an open-set domain generalization framework that extends ACGAN with dual-level weighting mechanisms to synthesize fault diagnosis data under unseen operating conditions. Similarly, \citep{jian2024gradient} introduces a gradient-based meta-learning strategy that augments the source domain to improve single-domain generalization. These works operate in the out-of-distribution (OOD) setting, where the objective is to generate or adapt to novel data distributions not present during training. In contrast, our work focuses on generating high-quality in-distribution data under data scarcity conditions. Rather than expanding beyond the original data distribution, we aim to faithfully reproduce it, particularly in critical domains such as healthcare and finance, where maintaining the statistical integrity of the data is essential.

Beyond GANs, Variational Autoencoders (VAEs) \citep{Kingma2013AutoEncodingVB} have emerged as a robust alternative for generating tabular data. Models such as TVAE \citep{xu2019modeling} are specifically designed to manage mixed data types, while HI-VAE \citep{nazabal2020handling} focuses on handling heterogeneous and incomplete datasets. VAEM \citep{ma2020vaem} extends this line by integrating imputation and generative modeling. More recently, the VAE-BGM \citep{apellaniz2024improved} has integrated a Bayesian Gaussian Mixture (BGM) in the VAE architecture, replacing the standard isotropic Gaussian assumption in the latent space \citep{Cremer2018InferenceSI}. This modification enables the model to sample from a more expressive latent space, generating higher-quality data, particularly in situations with limited availability. Additionally, emerging approaches based on diffusion models \citep{ho2020denoising} and transformer architectures have shown promise in generating synthetic tabular data. Models such as STaSy \citep{kim2022stasy} and TabSyn \citep{zhang2023mixed} apply score-based generative modeling with latent encoding strategies. In contrast, TabDDPM \citep{kotelnikov2023tabddpm} performs diffusion in both Gaussian and multinomial spaces to effectively model continuous and categorical features. These models benefit from improved training stability and are particularly effective in capturing complex inter-feature dependencies.

The importance of sufficient data for DGM training cannot be overstated. Studies using popular DGMs, such as CTGAN, utilize datasets ranging from $10,000$ to $481,000$ training instances. This contrasts starkly with the practices observed in numerous domains. For example, well-established datasets used to evaluate rudimentary models include the Iris dataset, with 150 samples \citep{misc_iris_53}, and the Boston House Prices dataset, with $506$ instances \citep{HARRISON197881}. Even within the realm of medical research, valuable datasets such as the breast cancer dataset encompass only around 300 patients \citep{misc_breast_cancer_14}. Smaller datasets pose challenges for DGM training, including overfitting and difficulty assessing the quality of generated data \citep{apellaniz2024synthetic}. 

A critical challenge associated with using DGMs for generating tabular data lies in ensuring the quality and effectiveness of the synthetic data. While standardized metrics exist for image \citep{heusel2018gans,salimans2016improved} and text data \citep{zhang2020bertscore, Papineni2002BleuAM}, measuring the quality of synthetic tabular data presents unique challenges. Studies employ various metrics, including pairwise correlation difference, support coverage, likelihood fitness, and other statistics described in \citep{vicomtech}. However, a consistent method for holistic evaluation is lacking. Divergences, which quantify the discrepancy between probability distributions, offer a promising avenue for validation \citep{apellaniz2024synthetic}. They can capture the overall differences between real and synthetic data by considering the joint distribution of all attributes. However, modeling joint distributions presents a trade-off between computational cost and accuracy. Large datasets, especially those with high dimensionality, require significant computational resources. With sufficient resources, accurate results can be achieved. In contrast, smaller datasets, common in real-world applications, present a challenge to accuracy. Limited data may hinder the capture of complex variable relationships, leading to models with poor generalization to unseen data. Consequently, even computationally efficient methods for joint distribution modeling can yield inaccurate results in small data settings.

The limitations of current validation techniques further compound the inherent limitations of small datasets. These techniques often focus on comparing synthetic data with real data used for training, failing to account for the limited scope of information on which the DGM was trained. This can lead to a false sense of security, where the synthetic data appear similar to the training data but may not generalize well. DL models with high parameter counts are susceptible to overfitting, too, especially on small datasets. Additionally, small datasets might not capture the full spectrum of real-world variations. Consequently, the synthetic data generated may not accurately represent the underlying distribution, which can affect its effectiveness for various tasks. Furthermore, smaller datasets are more prone to the influence of noise (random errors) and bias (systematic skews). These mislead DGMs into learning incorrect patterns, ultimately resulting in the generation of unrealistic synthetic data.

This work addresses the critical challenge of generating reliable synthetic tabular data from limited datasets, a prevalent scenario in many real-world applications. Traditional divergence metrics often struggle in these situations, leading to inaccurate assessments of the quality of synthetic data \citep{apellaniz2024synthetic}. We propose a novel methodology specifically designed to address this issue by introducing a framework that leverages inductive biases to improve the performance of DGMs in small dataset environments. Inductive biases \citep{goyal2022inductive} are inherent preferences or assumptions built into a learning model. These biases can guide learning and improve model performance, particularly when data are scarce. Traditionally, inductive biases are introduced through domain knowledge or specific architectural choices. This work proposes an alternative approach that leverages the variability often found in the DGM training process to generate inductive biases through different learning techniques. For example, we use VAEs, as they exhibit inherent variability between training seeds, allowing them to capture various aspects of data during training. Still, we note that our approach could be applied to other DGMs. We exploit this variability by employing various transfer learning and meta-learning techniques to generate the inductive bias, ultimately leading to improved synthetic data generation. Our key contribution is threefold:
\begin{itemize}
    \item We propose a novel generation methodology for synthetic tabular data generated by DGMs in a small dataset environment. This methodology leverages inductive bias generation through transfer learning and meta-learning techniques to achieve a more reliable generation process.
    \item We propose four different techniques (pre-training, model averaging, Model-Agnostic Meta-Learning (MAML), and Domain Randomized Search (DRS)) to generate the inductive bias. We demonstrate the efficacy of our proposal using a common DGM as the VAE. We also assess the performance with a CTGAN architecture, another common DGM for the pre-training process.
    \item We conduct extensive experiments on benchmark datasets and evaluate the quality of the generated data using both divergence-based metrics, such as the Kullback-Leibler (KL) and Jensen-Shannon (JS) divergences, and a utility-based validation protocol. This latter validation involves training a downstream model on synthetic data and testing it on real data, thereby measuring the practical utility of the generated samples for predictive tasks.
\end{itemize}

\section{Methodology: Generating Artificial Inductive Bias}\label{methods}
Assuming a tabular dataset composed of $N$ entries $\{x_{r}^i\}_{i=1}^N$, where $N$ represents the number of samples available and each entry $x_r^i$ has a dimensionality of $C$ features. In other words, $C$ represents the number of attributes associated with each data point. Let us also define a DGM as a high-dimensional probability distribution $p_{\theta}$, where $\theta$ represents the learnable parameters of the model. The objective of the DGM is to learn a representation, $p_{\theta}$, that closely approximates the true underlying data distribution, denoted by $p(x_r)$. Once trained, the DGM can generate new synthetic samples $x_{g}$ by drawing from its learned distribution:
\begin{equation}
    x_{g} \sim p_{\theta}.
\end{equation}
Ideally, a well-trained DGM should produce synthetic data $x_{g}$ that are statistically indistinguishable from real data $x_r$.

In the prevalent big data setting, characterized by a large number of training samples ($N \gg C$), DGMs with sufficient complexity can effectively capture the underlying data distribution $p(x_r)$. This is evidenced by the impressive results achieved in recent research, where high-dimensional synthetic samples are generated using vast amounts of training data \citep{xu2019modeling, elasri2022image, zhang2023survey, selva2023video}. However, for scenarios with limited training data, which is common in tabular domains, DGMs struggle to accurately represent the complex inter-feature relationships. Consequently, the synthetic samples generated $x_{g}$ deviate significantly from the true data distribution $p(x_r)$, leading to high KL and JS divergences between real and synthetic data.

\begin{figure}[!ht]
    \centering
    \includegraphics[width=\textwidth]{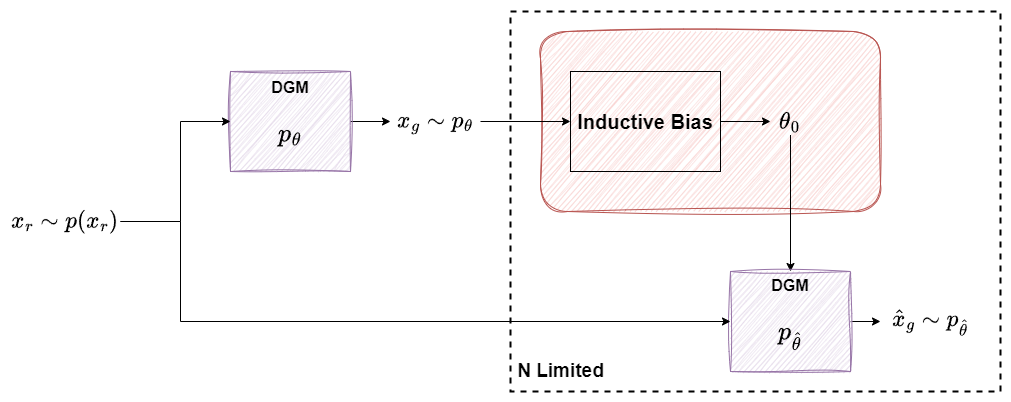}
    \caption{\textbf{Block diagram of the proposed architecture.} In standard big-data settings, the first generative model $p_{\theta}$ generates samples. However, when data are scarce ($N$ is limited), we use its output to create an artificial inductive bias, providing initial weights $\theta_0$ for a second DGM $p_{\hat{\theta}}$. This second model is then trained on real data $x_r$, producing higher-quality synthetic data $\hat{x}_{g}$ than the initial synthetic samples $x_{g}$.}

    \label{fig:general_flow}
\end{figure}

We propose an approach that uses artificially generated inductive biases to address this challenge. Fig. \ref{fig:general_flow} illustrates the overall architecture. In the standard big data setting, a DGM $p_\theta$ is directly trained using real data $x_r$, generating high-quality synthetic data $x_{g} \sim p_\theta$. However, when the number of real samples $N$ is limited, the quality of the generated data $x_{g}$ deteriorates. To mitigate this issue, we introduce an artificial inductive bias generator. This module inputs the initial synthetic data $x_{g}$ and outputs an initial set of weights $\theta_0$. These weights are then used as the inductive bias to train a second DGM $p_{\hat{\theta}}$ using real data $x_r$. This second DGM generates a new set of synthetic samples, $\hat{x}_{g}$. Notably, the only distinction between $p_{\theta}$ and $p_{\hat{\theta}}$ lies in the initial weights: $p_{\hat{\theta}}$ leverages the inductive bias encoded in $\theta_0$ to potentially achieve faster convergence to a distribution that better resembles $p(x_r)$. At the same time, $p_\theta$ begins training with random weights. As our simulations will demonstrate, this seemingly minor difference translates into significant improvements in the quality of the generated synthetic data.

The proposed approach hinges on two key concepts: the importance of inductive biases and the feasibility of generating them artificially. The importance of inductive biases in supervised learning is well established. The no-free-lunch theorems state that a universally optimal learner does not exist. Consequently, specific learning biases can lead to substantial performance gains for particular problem domains (see \cite{goldblum2023no} and the references therein). Convolutional Neural Networks (CNNs) exemplify this principle. Their inherent inductive bias, which is the fact that the image information possesses spatial correlation, makes them the preferred architecture for image processing tasks. Similarly, as highlighted in \citep{goyal2022inductive}, using inductive biases is a cornerstone of DL's success. In scenarios with limited training data, regularizers are commonly employed as inductive biases to prevent overfitting. This underscores the dual role of inductive biases: not only do they contribute to DL's effectiveness, but they are also crucial in avoiding overfitting. However, effective use of inductive biases is often contingent on having specific knowledge about the problem. In the aforementioned example of CNNs, we inherently understand spatial correlation in images. However, in tabular data, this domain-specific knowledge is often scarce. Recent efforts have focused on designing large models trained on artificially generated data as inductive biases to address this challenge. The underlying hope is that the actual problem to be solved exhibits similarities to those encountered during training of the large model (e.g., \cite{muller2021transformers} and \cite{hollmann2022tabpfn}).

Therefore, our proposed approach departs from existing methods for incorporating inductive biases in synthetic tabular data generation. Unlike the brute-force approach employed in \cite{hollmann2022tabpfn}, we use data generated by a potentially low-quality DGM $p_{\theta}$. This strategy aims to obtain an initial set of weights $\theta_0$, which act as an inductive bias. Ideally, these weights should guide the model to a region of the parameter space that facilitates convergence towards a high-quality solution. 

In particular, we assess our ideas using a state-of-the-art VAE architecture for the DGM, although we hypothesize that similar results could be achieved with other DGM architectures. Due to its demonstrated superiority against other leading models, we will use the architecture proposed in \cite{apellaniz2024improved}. VAEs are known to be sensitive to the initial random conditions (seeds) used during training. This dependence on seeds requires training with multiple seeds and selecting the one (or more) that exhibit the best performance based on a chosen metric, such as the minimum validation loss. The remaining runs, often discarded, may still contain valuable problem-specific information despite not achieving optimal solutions using traditional metrics. Our key idea lies in exploiting the potentially informative data from discarded VAE runs to create an artificial inductive bias for the final DGM, which is trained with real data. In addition to its demonstrated empirical performance, this VAE-based model incorporates architectural elements that inherently discourage overfitting and sample memorization, an important consideration in data-scarce regimes. Specifically, the model introduces a two-stage sampling process: first, latent representations are sampled from the encoder's posterior distribution; then, instead of decoding directly from these individual latent points, a BGM is fitted over the latent means, from which a second sample is drawn before decoding. This structure increases diversity in the generated samples and reduces the likelihood of memorizing specific training records. Moreover, the VAE formulation enforces a KL divergence regularization between the approximate posterior and the prior, further penalizing overly confident or localized representations that could lead to overfitting. These design choices serve as strong inductive biases that favor generalization over replication, complementing our overall strategy.

The following subsections explore two distinct paradigms for generating the initial set of weights, $\theta_0$: transfer learning and meta-learning. Transfer learning techniques encompass pre-training and model averaging, while meta-learning techniques include MAML and DRS. Pre-training offers a versatile approach applicable to any DGM architecture, regardless of its inherent characteristics. In contrast, model averaging and meta-learning techniques are particularly well-suited for VAEs trained with multiple seeds due to their intrinsic variability in learned representations. Consequently, we will evaluate the latter two methods within the chosen VAE architecture. Additionally, to assess the efficacy of pre-training across different DGM architectures, we will compare its performance on the CTGAN.

\subsection{Transfer learning}
Transfer learning is a machine learning paradigm that leverages knowledge acquired from a context domain (also called the source domain) to enhance learning performance in a new target domain \citep{Pan2010}. This approach aims to improve the learning process in the target domain by leveraging knowledge gained from solving related tasks in the context domain. This technique has demonstrated its efficacy in fields where data scarcity is a common challenge, such as the medical field \citep{Kim2022}.

Formally, based on the definition in \cite{Pan2010}, we can define a domain $\mathcal{D}$ by a feature space $\mathcal{X}$ and a marginal probability distribution $p(x_r)$. Two domains are considered distinct if their feature spaces $\mathcal{X}_1, \mathcal{X}_2$ or marginal probability distributions $p(x_1), p(x_2)$ differ, i.e., if $\mathcal{X}_1 \neq \mathcal{X}_2$ or $p(x_1) \neq p(x_2)$. The core objective of transfer learning is to leverage the knowledge learned in a context domain $\mathcal{D}_{context}$ to improve learning in a target domain $\mathcal{D}_{target}$. This is typically achieved when the context and target domains differ, i.e., $\mathcal{D}_{context} \neq \mathcal{D}_{target}$. 

Our work focuses on a scenario where the context domain $\mathcal{D}_{context}$ consists of data $x_g$ generated by a DGM. On the other hand, the target domain $\mathcal{D}_{target}$ consists of $x_r$. Our approach leverages the representational power learned by the DGM $p_\theta$ on $x_g$ to provide a strong starting point for learning in the target domain with real data $x_r$. This knowledge transfer is achieved by initializing the model weights for the target domain with the weights learned from the DGM model trained on the generated data.

Transfer learning can be categorized into homogeneous and heterogeneous settings based on the feature spaces of the domains \citep{weiss2016survey}. Homogeneous transfer learning applies when the context and target domains share the same feature space $\mathcal{X}_1= \mathcal{X}_2$, while heterogeneous transfer learning deals with scenarios where feature spaces differ $\mathcal{X}_1\neq \mathcal{X}_2$. This work focuses on homogeneous transfer learning, where the context domain is an augmented version of the target domain. The key difference between the domains in our case lies in the number of samples, leading to situations where the empirical distributions of the data differ, i.e., $p_\theta(x_g) \neq p(x_r)$.

Within homogeneous transfer learning, various methodologies exist to improve target task performance by leveraging knowledge from a related source domain. These techniques encompass instance-based \citep{Chawla_2002}, relational knowledge transfer \citep{li-etal-2012-cross}, feature-based \citep{Long2014AdaptationRA}, and, as employed in this work, parameter-based \citep{Duan2012ExploitingWI} transfer through shared model parameters or hyperparameter distributions. This study leverages a two-stage parameter-based transfer learning approach. The first stage involves pre-training or model averaging, followed by fine-tuning in the second stage. Subsequent sections will delve deeper into both pre-training and model averaging techniques. Upon completing one of these initial phases, fine-tuning refines the model parameters, ultimately achieving optimal adaptation for the target domain.

\subsubsection{Pre-training}
Pre-training is a frequently adopted strategy for introducing an inductive bias into a model. By leveraging a pre-trained model on a context domain, the target model gains generalizable features that enhance its performance on a target domain. However, while pre-training is a standard in computer vision and natural language processing, achieving similar success with tabular data remains challenging. This disparity arises from the inherent heterogeneity of the features in the tables, which creates substantial shifts in feature space between pre-training and downstream datasets, hindering effective knowledge transfer. Despite these challenges, efforts such as \cite{wang2022transtab} and \cite{zhu2023xtab} have explored tabular transfer learning with promising results. Although these studies demonstrate potential, achieving comprehensive parameter transfer in tabular data requires further research to establish best practices and fully leverage the potential of pre-training in this domain.

In this work, pre-training involves the following steps. First, we train a separate DGM $p_{\theta_{pt}}$ using synthetic data $x_{g}$ as training data. Since $x_{g}$ is sampled from the initial DGM, $x_{g} \sim p_{\theta}$, we can generate a vast amount of synthetic data. This abundance circumvents the limitations associated with training in small datasets, such as overfitting. Then, the optimal weights $\theta_{pt}^*$  from DGM $p_{\theta_{pt}}$ are used as initial weights $\theta_0$ to train the generative model $p_{\hat{\theta}}$  (see Fig. \ref{fig:general_flow}).

\begin{figure}[!ht]
    \centering
    \includegraphics[width=\textwidth]{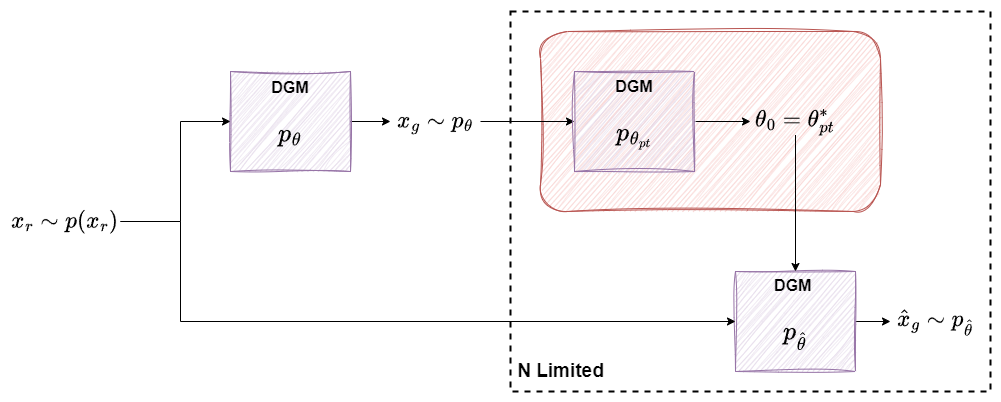}
    \caption{\textbf{Block diagram for the pre-training case.} The inductive bias is introduced by training a DGM $p_{\theta_{pt}}$ on a large collection of $x_{g}$ samples. The weights learned from this training process with abundant samples serve as the initial parameters $\hat{\theta}$ for the fine-tuning process using the real data $x_r$ to obtain $p_{\hat{\theta}}$.}
    \label{fig:flow_pre_train}
\end{figure}

In essence, our approach aligns with the well-established concept of data augmentation. We generate synthetic data $x_{g}$, which may not perfectly capture the intricacies of the original data $x_r$. However, we used these synthetic data to train another DGM $p_{\theta_{pt}}$. Although $p_{\theta_{pt}}$ might generate lower-quality synthetic samples, our objective is to exploit the information encoded within this DGM to establish an initial set of weights for the DGM that will eventually be trained on $x_r$. In other words, we exploit the knowledge of the generative model $p_{\theta_{pt}}$, the context domain, to obtain a better generative model $p_{\hat{\theta}}$, which is our target domain. Fig. \ref{fig:flow_pre_train} visually represents this pre-training procedure.

\subsubsection{Model Averaging}
The concept of model averaging emerged in the 1960s, primarily within the field of economics \citep{model_averaging, d890a735-c271-388b-8b71-a309e52af62b}. Traditional empirical research often selects a single ``best" model after searching a wide space of possibilities. However, this approach can underestimate the real uncertainty, leading to overly confident conclusions. Model averaging offers a compelling alternative. By combining multiple models, the resulting ensemble can outperform any individual model. This approach aligns with the core principles of statistical modeling, which involve maximizing information use while balancing flexibility with the risk of overfitting. In essence, model averaging extends the concept of model selection by leveraging insights from all the models considered.

While pre-training can be incorporated with any DGM, our approach focuses on models where the training process is sensitive to initial conditions, such as VAEs. In such cases, it is common to train the DGM $p_\theta$ with multiple initial conditions (seeds) and potentially discard ``bad" seeds based on a specific metric. We propose using these discarded seeds to create an artificial inductive bias. The simplest implementation involves averaging the model parameters.
In this case, our context domains are the different results obtained from each seed, and the target domain is obtained by averaging across these context domains. If we train $S$ different seeds for $p_\theta$, resulting in $S$ models with parameters $\theta_s$, we propose using the average of these weights as the inductive bias:
\begin{equation}
    \theta_0 = \frac{1}{S} \sum_{s=1}^S \theta_s
\end{equation}

\begin{figure}[!ht]
    \centering
    \includegraphics[width=\textwidth]{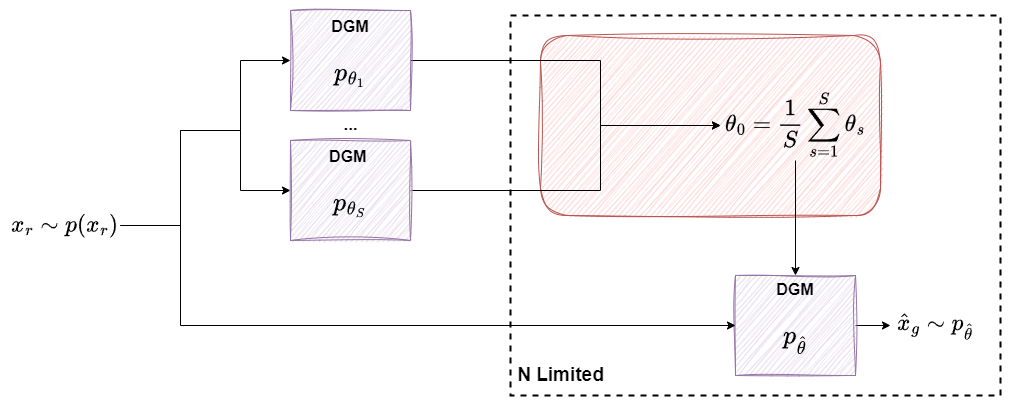}
    \caption{\textbf{Block diagram for the model averaging case.} The inductive bias is introduced by training a DGM $p_{\theta}$ on $x_r$ using $S$ different seeds. The average of the weights learned from these training processes serves as the initial parameters $\hat{\theta}$ for the fine-tuning process using the real data $x_r$ to obtain $p_{\hat{\theta}}$.}
    \label{fig:flow_mod_avg}
\end{figure}

This straightforward approach is computationally efficient, requiring only the calculation of the average across the precomputed weights. It assumes that the average model may capture a robust inductive bias, leading to improved performance. Fig. \ref{fig:flow_mod_avg} summarizes this process.

\subsection{Meta-learning}
Traditional machine learning models often rely on large volumes of data to achieve optimal performance in specific tasks. In contrast, meta-learning introduces a distinct paradigm by training algorithms that can ``learn to learn" \citep{Thrun1998}, enabling them to rapidly adapt to new tasks with minimal data. This departure from the conventional requirement of extensive datasets for each new task allows meta-learning algorithms to leverage knowledge by addressing numerous related tasks. Through introspective analysis of past experiences, these models dynamically adjust their learning strategies when confronted with novel situations, making them more efficient learners and requiring less data to perform well on tasks with similar characteristics. 

In this work, we exploit the multi-seed training configuration of certain DGMs. We construct a meta-learning framework by treating each $S$ different seeds obtained after training the DGM as a distinct task. 

\subsubsection{MAML}
MAML is a prevalent approach within the field of meta-learning \citep{finn2017model}. It identifies the initial set of weights denoted by $\theta_{MAML}$ by leveraging various tasks, enabling rapid and data-efficient adaptation to new tasks. This efficiency comes from fine-tuning the $\theta_{MAML}$ with minimal data for each new task. However, the successful application of MAML requires access to diverse tasks for effective learning.

Formally, we can frame the problem by starting with a common single-task learning scenario and transforming it into a meta-learning framework. Consider a task $\mathcal{T}$ that consists of an input $x$ sampled from a probability distribution $\mathcal{D}$. For simplicity, we define a task instance $\mathcal{T}$ as a tuple comprising a dataset $\mathcal{D}$ and its corresponding loss function $\mathcal{L}$. To solve the task $\mathcal{T}$, we need to obtain an optimal model parameterized by a task-specific parameter $\omega^*$, which minimizes a loss function $\mathcal{L}$ on the data of the task as follows:
\begin{equation}
    \omega^* = \arg \min_{\omega}\underset{x \sim \mathcal{D}}{\mathbb{E}} \big[\mathcal{L}(\mathcal{D}; \omega)\big].
\end{equation}

In single-task learning, hyperparameter optimization is achieved by splitting the dataset $\mathcal{D}$ into two disjoint subsets $\mathcal{D} = \mathcal{D}^{(t)} \cup \mathcal{D}^{(v)}$, which are the training and validation sets, respectively. The meta-learning setting aims to develop a general-purpose learning algorithm that excels across a distribution of tasks represented by $p(\mathcal{T})$ \citep{Hospedales2020MetaLearningIN}. The objective is to use training tasks to train a meta-learning model $\theta_{MAML}$ that can be fine-tuned to obtain $\omega$ to perform well on unseen tasks sampled from the same task environment $p(\mathcal{T})$. Meta-learning methods utilize meta-parameters to model the common latent structure of the task distribution $p(\mathcal{T})$. Therefore, we consider meta-learning an extension of hyperparameter optimization, where the hyperparameter of interest – often referred to as a meta-parameter – is shared across multiple tasks. 

In this work, the distribution of tasks is defined by the set of $S$ training seeds obtained after training the DGM. Given a set of $S$ training seeds following $p(\mathcal{T})$, each task $\mathcal{T} \sim p(\mathcal{T})$ is therefore formalized as $\mathcal{T} = \{\mathcal{D}, \mathcal{L}\}$. Each dataset $\mathcal{D}$ consists of synthetic data points $x_{g}^s$ drawn from the model for the different training seeds. The loss function $\mathcal{L}$ corresponds to the DGM loss function. The specific $\mathcal{L}$ form depends on the chosen DGM. If the chosen DGM is a VAE, the loss function $\mathcal{L}$ would be the negative of the Evidence Lower BOund (ELBO) \citep{Kingma2013AutoEncodingVB}. In contrast, if a GAN is used, the loss function $\mathcal{L}$ would be the minimax loss function arising from the interplay between the generator and discriminator networks \citep{Goodfellow2014GenerativeAN}. It is essential to note that both VAEs and GANs employ two neural networks within their architecture, which differs from the single network architectures commonly found in state-of-the-art applications \citep{Gordon2018MetaLearningPI, SmithMiles2009CrossdisciplinaryPO}.

\begin{figure}[!h]
    \centering
    \includegraphics[width=\textwidth]{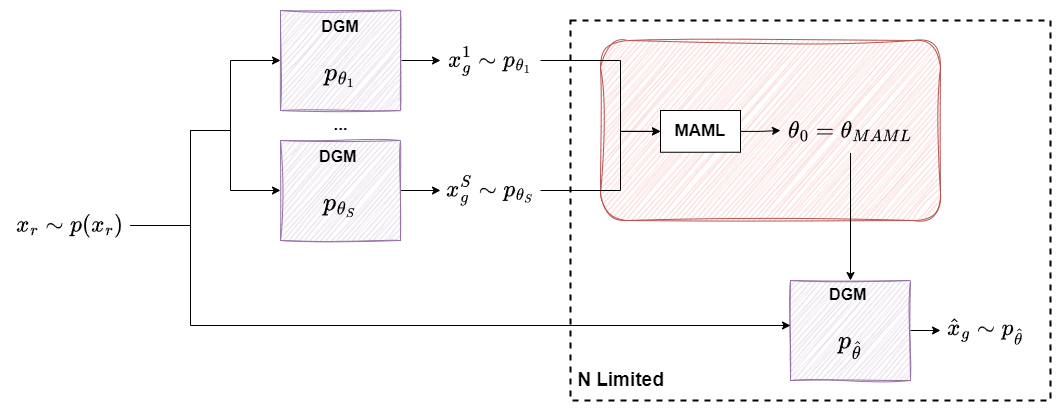}
    \caption{\textbf{Block diagram for the MAML case.} The inductive bias is introduced by training a DGM $p_{\theta}$ on $x_r$ using $S$ different seeds. The synthetic dataset $x_{g}^s$ generated by each seed serves as a task for MAML. The starting point to fine-tune using the real data $x_r$ and obtain $p_{\hat{\theta}}$ is the MAML solution obtained, $\theta_{MAML}$.}
    \label{fig:flow_maml}
\end{figure}

Solving this problem using the MAML approach requires access to a collection of $B$ tasks sampled from $p(\mathcal{T})$. We denote this set of tasks $\mathcal{T}_b$ used for training as $\mathcal{D}_{b} = \{(\mathcal{D}_{b}^{(t)}, \mathcal{D}^{(v)}_b)\}_{b=1}^B$, where each task $b$ has dedicated meta-training and meta-validation data, respectively. The goal of meta-training is to find the optimal $\omega^{*}_b$ for a given task $b$ given $\theta_{MAML}$. This $\theta_{MAML}$ essentially captures the ability to learn effectively from new data. In this context, the task-related parameter $\omega_b$ refers to the task-specific parameters of the two networks that comprise the VAE, namely, the encoder and decoder. After meta-training, the learned $\omega^{*}_b$ is used to guide the training of a base model $\theta_{MAML}$. This procedure is called meta-testing. This essentially means that the model leverages the knowledge gained from previous tasks to improve the efficiency of learning on new tasks. This can be viewed as a bi-level optimization problem \citep{Franceschi2018BilevelPF}:
\begin{equation}\label{eq:inner_outer}
\begin{split}
    \min_{\theta_{MAML}} \underset{\mathcal{T}_b \sim p(\mathcal{T})}{\mathbb{E}} \left[\underset{x_{g_b}^{(v)} \sim \mathcal{D}^{(v)}_b}{\mathbb{E}} \big[\mathcal{L}_b(\mathcal{D}^{(v)}_b; \omega^{*}_b(\theta_{MAML}))\big]\right] \\
    \textrm{s.t: }\omega^{*}_b(\theta_{MAML}) = \arg \min_{\omega_b} \underset{x_{g_b}^{(t)} \sim \mathcal{D}^{(t)}_b}{\mathbb{E}} \big[\mathcal{L}_b(\mathcal{D}^{(t)}_b; \omega_b(\theta_{MAML})) \big].
    \end{split}
\end{equation}

 This equation minimizes the expected loss across all tasks on the meta-validation sets, subject to the constraint that the task-specific parameter $\omega$ is optimized on the corresponding meta-training data for each task.

Since in our work, we are upgrading the parameters using gradient descent, we can reformulate \Cref{eq:inner_outer} as follows:
\begin{align}
    &\omega_b \leftarrow \theta - \alpha \nabla_{\omega_b}\mathcal{L}_b(\mathcal{D}^{(t)}_b;\omega_b) \\
    &\theta_{MAML} \leftarrow \theta_{MAML} - \gamma \nabla_{\theta_{MAML}}\sum_{b=1}^B\mathcal{L}_b(\mathcal{D}^{(v)}_b;\theta_{MAML}).
\end{align}

Here, $\alpha$ and $\gamma$ represent the learning rates for the inner and outer loops, respectively. The inner loop updates the task-specific parameters $\omega$ for each task $b$ using the gradient of the loss function $\mathcal{L}_b$ in the meta-training data. The outer loop updates the meta-parameters $\theta_{MAML}$ based on the accumulated meta-validation loss across all tasks.

Fig. \ref{fig:flow_maml} illustrates the integration of the MAML procedure within the framework of our proposed methodology. In this context, the task space denoted by $p(\mathcal{T})$ corresponds to the various seeds $S$ obtained during the training process. Essentially, the task space encompasses the different probability distributions $p_{\theta_s}$ associated with each training seed. Ultimately, the meta-training steps lead to identifying the desired parameters, denoted by $\theta_{MAML}$. Note that $\theta_{MAML}$ represents a set of parameters that adapt fast to new data; in our case, it means that the DGM initial parameters are chosen so that they adapt fast to generate real data.

\subsubsection{DRS}
Although MAML offers the potential to leverage the underlying structure of learning problems through a powerful optimization framework, it introduces a significant computational cost. Therefore, while we should seek a trade-off between accuracy and computational efficiency, there is no approach to managing this trade-off. An understanding of the domain-specific characteristics inherent to the meta-problem itself is needed.

DRS presents an alternative meta-learning approach that circumvents the computational burden of bilevel optimization problems. Unlike MAML, DRS trains a model on the combined data from all tasks. This eliminates the need for the complex optimization procedures present in MAML, leading to a more computationally efficient solution. However, it is important to acknowledge that DRS approximates the ideal solution \citep{gao2020modeling}.

Formally, DRS focuses on the meta-information, denoted by $\theta_{meta}$, as the initialization of an iterative optimizer used in a new meta-testing task, $\mathcal{T}_S$. In this context of meta-learning initialization, a straightforward alternative involves solving the following pseudo-meta problem:
\begin{equation}
    \theta_{DRS} = \arg \min_{\omega}  \underset{\mathcal{T_S}\sim p(\mathcal{T})}{\mathbb{E}}\mathcal{L}(\mathcal{D^*}; \omega) .
\end{equation}

In this context, $\mathcal{D^*}$ represents the aggregated synthetic data collection, $x_{g}$, obtained across all training seeds $S$. We refer to this approach as Domain-Randomized Search due to its alignment with the domain randomization method presented in \cite{tobin2017domain} and its core principle of directly searching over a distribution of domains (tasks).

\begin{figure}[!ht]
	\centering
        \includegraphics[width=\textwidth]{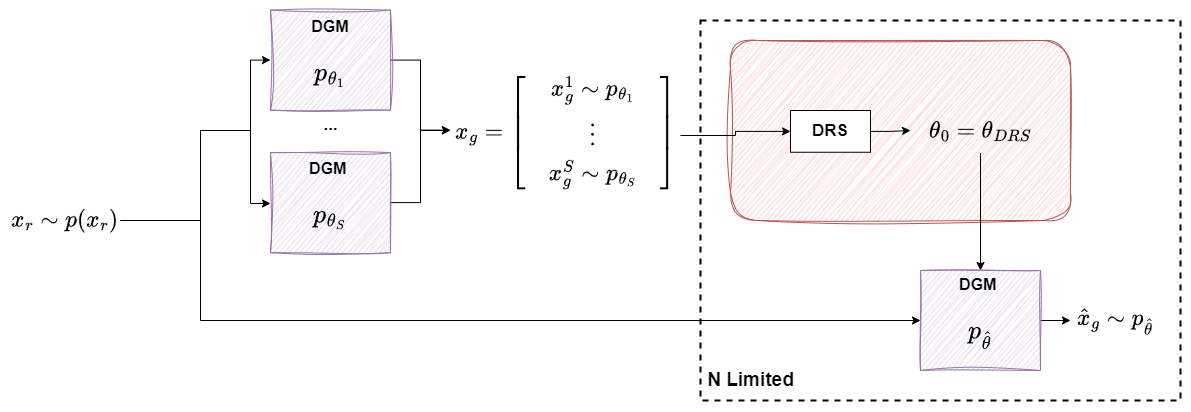}
        \caption{\textbf{Block diagram for the DRS case.} The inductive bias is introduced by training a DGM $p_{\theta}$ on $x_r$ using $S$ different seeds. The synthetic dataset $x_{g}$ contains data generated by each seed and serves as input to DRS. The starting point for fine-tuning using the real data $x_r$ and obtaining $p_{\hat{\theta}}$ is the DRS solution, $\theta_{DRS}$.}
	\label{fig:flow_drs}
\end{figure}

Fig. \ref{fig:flow_drs} shows the application of the DRS procedure within the framework of our proposed methodology. Like the MAML case, $\theta_{DRS}$ is the initialization weights we aim to identify $\theta_0$.

Both MAML and DRS offer complementary approaches with a trade-off between modeling complexity and optimization cost \citep{gao2020modeling}. DRS delivers an approximate solution with lower computational demands, while MAML offers higher precision at the expense of greater computational resources. `DRS' is also advantageous when dealing with a limited number of learning tasks. In our case, where data generated by each seed ($s=1,2,..., S$) is considered a task, and $S$ typically takes values around $10$, DRS is expected to provide better solutions than MAML, aligning with the findings of \cite{gao2020modeling}. Finally, note that DRS is similar to the pre-training approach. While both techniques aim to improve model performance, they utilize data differently. Pre-training leverages data from the best VAE seed, whereas DRS capitalizes on data from all VAE seeds. This distinction reflects the core principle of DRS: exploring a wider range of possibilities by searching across a distribution of domains (tasks) represented by the various seed variations.

\subsection{Summary and Comparison of Techniques}
To clarify the overall methodology, we provide a high-level step-by-step summary of the artificial inductive bias generation process. (1) First, a DGM is trained using the limited real data $x_r$ or large synthetic data $x_g$ (sampled from a previously trained model), depending on the technique. (2) Then, an artificial inductive bias is constructed by either extracting weights (pre-training), averaging multiple trained models (model averaging), or applying a meta-learning algorithm across multiple training seeds (MAML or DRS). (3) These bias-derived weights $\theta_0$ are then used to initialize a second DGM, which is finally fine-tuned on the real data $x_r$ to obtain improved synthetic samples $\hat{x}_g$. This two-phase process enables the model to start learning from a more informative point in the parameter space, helping to mitigate overfitting or collapse in low-data regimes.

Each technique integrates this framework differently. Pre-training mitigates data scarcity by simulating a high-data regime with synthetically generated samples, while model averaging consolidates information from multiple VAE training seeds to create a more robust initialization. MAML optimizes an initialization that adapts quickly to new tasks, using the diversity of seed-generated data as a proxy for task variability. DRS simplifies this by combining all seeds into a single learning task, offering a trade-off between accuracy and computational burden. While pre-training is flexible and compatible with any architecture, it may transfer suboptimal representations if synthetic data quality is poor. Model averaging is simple and effective for VAEs, but it assumes that the mean of the parameter space corresponds to a good initialization. MAML can yield precise adaptations but may require more task diversity than available in low-seed settings. DRS is more robust under these conditions but offers a less principled optimization. Our approach addresses these limitations by leveraging the strengths of each technique within the VAE framework and evaluating them under consistent conditions. Limitations such as the architecture dependence of certain techniques and the impact of seed variability are discussed further in the results and conclusions.

\section{Experiments}\label{results}
\subsection{Data} \label{sec:data}
The experiments were conducted on three public datasets, one from the SDV environment \citep{sdv}--which also provides the CTGAN implementation used in this study--and the other two additional datasets from external repositories. We selected datasets with sufficient samples to allow multiple training and validation splits, ensuring a comprehensive evaluation of our method under different parameter settings.

\begin{itemize}
    \item \textbf{Adult:} The Adult Census Income dataset \citep{adult} contains 32,561 samples with 14 mixed-type features (integer, categorical, binary). It is used to predict whether an individual's annual income exceeds $\$50,000$. The dataset has 13\% missing values, concentrated in ``work class" and ``occupation."
    
    \item \textbf{King\footnote{Source: \url{https://www.kaggle.com/datasets/harlfoxem/housesalesprediction} Accessed January $30^{\text{th}}$, 2025}:} The King County House Sales dataset contains 21,613 house sale records from King County, Washington, including Seattle, between May 2014 and May 2015. It includes 20 numerical and categorical features relevant to house pricing.
     
    \item \textbf{Letter:} The Letter Recognition dataset \citep{letter_recognition_59} includes 20,000 samples and 16 numerical features extracted from raster images of capital letters. The task is to classify each sample into one of 26 possible alphabet letters, making it a challenging multiclass classification problem. This dataset was obtained from the UCI Machine Learning Repository\footnote{Source: \url{https://archive.ics.uci.edu/dataset/59/letter+recognition} Accessed June $17^{\text{th}}$, 2025}.
   
\end{itemize}

This selection of datasets with varying sample sizes, feature dimensions, and task types (regression, binary, and multiclass classification) allows us to assess the performance and generalizability of the proposed method in diverse data scenarios. As will be demonstrated in the following sections, the proposed method is effective in generating synthetic data that retain the key characteristics of the original datasets across a wide range of applications. Additional datasets used for extended validation are described in \ref{app:additional_data}.

\subsection{Validation metrics}
To evaluate the effectiveness of our proposed method in capturing the real data distribution, we follow the validation approach from \cite{apellaniz2024synthetic}. This method uses a probabilistic classifier (discriminator) to estimate the ratio of probability densities between real and synthetic data, enabling the calculation of KL and JS divergences. Unlike traditional validation methods that focus on individual features, this approach considers the entire data distribution, including complex relationships between features. Divergences provide robustness to noise and clear interpretability, making them ideal for assessing the similarity between real data $p(x_r)$ and synthetic data $p_\theta$.

The discriminator network is key to the validation process, trained to distinguish between real and synthetic data. It receives two sets of $M$ samples: one from the real data distribution $p(x_r)$ (labeled as class 1) and another from the synthetic data distribution $p_\theta$ or $p_{\hat{\theta}}$ (labeled as class 0), depending on $N$, the dataset size. The discriminator learns a decision boundary during training, capturing differences between real and synthetic distributions. Once trained, it estimates KL and JS divergences by processing $L$ new samples from each distribution. The output probabilities of these samples are then used to compute the divergence metrics, providing a quantitative measure of distributional similarity.

\begin{figure}[!ht]
	\centering
        \includegraphics[width=\textwidth]{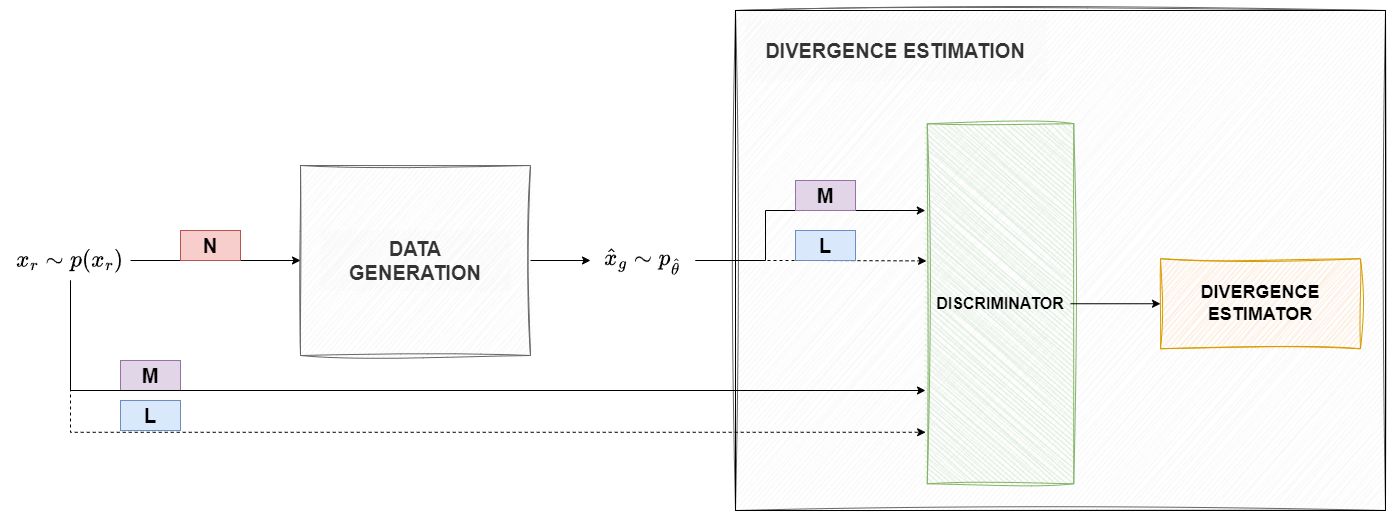}
        \caption{\textbf{General Scheme of the proposed approach.} Overall scheme of the approach and its validation process from \cite{apellaniz2024synthetic}. This last one consists of a discriminator and a divergence estimator. The number of samples used from each distribution for each step is also highlighted: $N$ to generate samples, $M$ to train the discriminator, and $L$ to estimate the divergences.}
	\label{fig:valid_scheme}
\end{figure}

Fig. \ref{fig:valid_scheme} illustrates this process, highlighting the inductive bias generator, discriminator-based validation, and the different sample sizes used ($N$ for generation, $M$ for training, and $L$ for divergence estimation). As stated in \cite{apellaniz2024synthetic}, $M$ and $L$ must be large enough to ensure reliable divergence estimates, which we consider in our experiments.

In addition to this divergence-based evaluation, we incorporate a utility-based validation protocol (also referred to as machine learning efficacy in the literature (e.g., \cite{xu2019modeling}). This approach assesses the utility of synthetic data in downstream predictive tasks by training machine learning models with synthetic samples and evaluating their performance on real data. Specifically, we define two configurations:

\begin{itemize}
    \item \textbf{Real \textit{metric}}: The model is trained and validated using real data. This provides an upper-bound reference for predictive performance.
    \item \textbf{Synthetic \textit{metric}}: The model is trained on synthetic data and evaluated on real data. This reflects how well the synthetic data preserve task-relevant structure.
\end{itemize}

Higher classification metrics or lower regression errors in the Synthetic configuration indicate that the generated data maintain high utility for downstream tasks. In classification tasks, the utility \textit{metric} corresponds to accuracy, while in regression tasks it corresponds to the MSE. Together, both divergence-based and utility-based evaluations provide a more comprehensive view of the quality and applicability of the synthetic data.

\subsection{Experimental design}
To evaluate the proposed methodology, we designed a comprehensive experimental pipeline including both divergence-based and utility-based validation. As described in the methodology section, we use the state-of-the-art VAE architecture \citep{apellaniz2024improved} and train 10 different seeds per experiment to account for stochastic variability and to enable the application of meta-learning and averaging techniques. We also evaluate the pre-training strategy using CTGAN, a well-established conditional GAN model for generating tabular data.

The VAE used in our experiments is based on a VAE with a BGM \citep{apellaniz2024improved}, which replaces the standard Gaussian latent assumption with a learned mixture model, enabling more flexible and expressive latent representations. Both encoder and decoder are fully connected neural networks with ReLU activations, a hidden layer of $256$ neurons, and dataset-specific latent dimensions: $10$ for Adult, Intrusion, Letter, and PenDigits; $15$ for King; and $20$ for News. The model is trained using the standard ELBO loss, which includes a KL divergence regularizer and employs early stopping to prevent overfitting. All training is conducted on the CPU with a batch size of $256$ and a default learning rate of $1 \times 10^{-3}$.

The CTGAN model \citep{xu2019modeling} is implemented using the SDV library’s default configuration \citep{sdv}, with $1,000$ training epochs and GPU acceleration enabled. It employs a conditional generator and a discriminator trained adversarially and includes a conditional vector strategy to handle categorical imbalance and mode-specific normalization for numerical features.

We use 5-fold cross-validation throughout all experiments, consistently splitting the real data into training and validation folds. For both divergence and utility validations, we define two configurations: a reliable setting with $M = 7500$ and $L = 1000$ samples and a more realistic low-resource setting with $M = L = 100$. The sample size $N$ for training the generative model is fixed based on the scenario being simulated and is not manipulated.

Utility-based evaluation is conducted across classification and regression tasks, depending on the dataset. In each case:
\begin{itemize}
    \item \textbf{Classification} uses a custom neural network with three hidden layers (256–64–32), batch normalization, dropout, and LeakyReLU activations. Training runs for $10,000$ epochs with early stopping.
    \item \textbf{Regression} employs a single-layer linear model trained for $2,000$ epochs using Mean Squared Error (MSE) loss and Adam optimizer.
\end{itemize}

The experimental results are grouped into the following scenarios:

\begin{enumerate}
    \item \textbf{`Big data':} First, we present the optimistic scenario with a sufficient $N$ of $10,000$ samples, where no methodology is needed to calculate the inductive bias. While we acknowledge that the term `big data' is typically associated with datasets containing hundreds of thousands or even millions of records, in this experimental design, we use it in a relative sense to contrast this setting with the data-scarce scenarios that are the primary focus of our study. This configuration provides divergence results that serve as an ``upper bound" or reference for the best possible outcome. For $M$ and $L$, we maintain high values of the validation samples, $7500$ and $1000$, respectively.
    \item \textbf{`Low data':} Next, we show results for a realistic scenario with few samples ($N=300$) without applying our methodology. This allows us to quantify the gain using the method and determine its benefits. For this case, we use two configurations for parameters $M$ and $L$: [$M=7500$, $L=1000$] and [$M=100$, $L=100$]. The second configuration is more realistic for few-data scenarios. When limited training data are available, there is also a limitation on the amount of data that can be effectively used for validation. The first configuration, with much larger values for $M$ and $L$, serves as a rigorous evaluation of the impact of our methodology. However, it is acknowledged that using small values for $M$ and $L$ can lead to unreliable metric estimations \citep{apellaniz2024synthetic}.
    \item \textbf{`Pre-train':} In this case, we apply the proposed methodology using the pre-training technique. Results are presented for both CTGAN and VAE. The parameter configurations chosen are: [$N=300$, $M=7500$, $L=1000$] and [$N=300$, $M=100$, $L=100$].
    \item \textbf{`AVG,' `MAML,' `DRS':} These scenarios apply the model average (`AVG') and the meta-learning techniques (`MAML,' `DRS'). We solely utilize the VAE architecture for multiple training runs. The following parameter configurations will be presented: [$N=300$, $M=7500$, $L=1000$] and [$N=300$, $M=100$, $L=100$].
\end{enumerate}

This setup aims to thoroughly evaluate the performance and robustness of the proposed methodologies in various data availability scenarios and parameter configurations. Hardware and software specifications are detailed in \ref{app:times}.

\subsection{Results}
In this section, we present the experimental results, focusing on scenarios with higher values of $M$ and $L$, which promote more reliable divergence estimation. Additional results for other datasets and the low-resource configuration ($M=100$, $L=100$) are provided in \ref{sec:app_exp} for completeness. For each database, we summarize the outcomes across all scenarios using both divergence-based and utility-based validation protocols. Divergence metrics include JS and KL divergences. Utility metrics reflect the performance of a downstream model trained with synthetic data. Specifically, we consider two setups: training and validating with real data (Real \textit{metric}) and training with synthetic data and validating on real data (Synth \textit{metric}). For classification tasks, the metric used is accuracy; for regression tasks, the metric used is the MSE. All results are reported as mean (standard deviation) across 10 independent runs with different random seeds. 

The code to replicate our results, along with the data used, can be found in our \href{https://github.com/Patricia-A-Apellaniz/low_sample_data_generator}{\textit{repository}}. Additionally, \ref{app:times} presents a comparative summary of the computational costs associated with each method, including training times and hardware and software configurations. This analysis supports practitioners in evaluating trade-offs between performance and resource requirements when selecting a technique.

Subtable (a) in \Cref{table:data_res} shows the validation results obtained for the Adult dataset. Regarding the divergence estimations, we focus primarily on the JS divergence due to its interpretability as a bounded metric (ranging from 0 to 1). The table shows the upper and lower bounds used to assess the efficacy of the proposed methodology in the reliable case of higher validation samples ($M=7500$ and $L=1000$). These bounds are $0.115$ (upper) and $0.420$ (lower), highlighting a significant gap and room for improvement in the base VAE model (without any techniques applied). A consistent decrease in divergence is observed when examining the JS divergence results for the application of different proposed techniques. The worst improvement is obtained for `MAML' ($0.358$) and the best for `AVG' ($0.163$). This implies that improvement is always present and, in the best cases, significantly high in JS for VAE. A similar pattern is observed for KL divergence in VAE: better divergence results are obtained for transfer learning cases, but improvements are always achieved. For the CTGAN model (where only pre-training results are available), we also observe a significant improvement in both JS and KL divergence despite the generation process appearing to be less accurate in terms of divergence estimation compared to the VAE. Regarding the utility validation, training with synthetic data generated by the VAE in the `Big data' scenario yields accuracy levels close to those obtained with real data. In the `Low data' case, the accuracy drops when using synthetic data but increases again when any of the proposed techniques are applied, validating the utility of these strategies. However, no improvement is observed in utility metrics when applying the CTGAN-based pre-training, suggesting that while divergence metrics may improve, they do not necessarily translate into enhanced downstream performance for this model.

Subtable (b) in \Cref{table:data_res} further reinforces the efficacy of the proposed methodologies on the King dataset. The VAE model consistently improves on the lower bounds established for the JS divergence across all techniques. In terms of KL divergence, the most notable improvement is observed with the `AVG' technique, which significantly reduces the KL value from $5.963$ to $3.322$. In contrast, CTGAN results are less favorable, with divergences that remain close to or above the baseline (`Low data') and offer little gain in the presence of pre-training. These results suggest that CTGAN is less robust in modeling the complex structure of regression data in small-sample regimes. The utility validation, as measured by MSE, mirrors the trends observed in the divergence results for both generative models. The synthetic data generated by the VAE under `Low data' conditions shows a substantial increase in MSE compared to the real-data baseline, indicating reduced predictive utility. However, applying techniques such as `AVG,' `Pre-train,' or `DRS' significantly reduces the MSE, bringing it closer to the values obtained in the `Big data' scenario. This validates the practical benefits of introducing inductive biases even in regression contexts. The technique `MAML,' in this case, appears less effective in terms of utility, likely due to the higher sensitivity of regression tasks to noise and instability during meta-learning optimization.

Ultimately, subtable (c) in \Cref{table:data_res} presents the results obtained for the Letter dataset, which involves a challenging multiclass classification task with 26 output classes. Regarding the divergence metrics, the application of the proposed techniques consistently improves JS divergence for the VAE model, except for `MAML,' which yields no improvement over the baseline scenario. A similar trend is observed in KL divergence: while most techniques achieve a notable reduction, `MAML' again fails to improve over the baseline, possibly due to optimization instability or suboptimal adaptation in this high-class-cardinality setting. For the CTGAN model, we observe no improvement in JS divergence but a significant reduction in KL divergence. In terms of utility validation, measured via accuracy, the only meaningful improvement over the `Low data' baseline is observed with the `DRS' technique. All other methods yield marginal gains or similar performance, indicating that despite improvements in distributional similarity, they do not translate into enhanced classification performance for this dataset. This may be due to the complexity of the classification task or the need for more targeted biasing strategies to capture high-granularity class structures better.

\begin{table}[!ht]
\renewcommand{\arraystretch}{1.1}
\centering

\begin{subtable}{\linewidth}\centering
{
\resizebox{\textwidth}{!}{%
\begin{tabular}{l|c|cccc|ccc}
\hline
\hline

\multicolumn{1}{l|}{\multirow{2}{*}{\textbf{Scenario}}} & 
\multicolumn{1}{c|}{\multirow{2}{*}{\textbf{N}}} &
\multicolumn{4}{c|}{\textbf{DIVERGENCE VALIDATION}}   & 
\multicolumn{3}{c}{\textbf{UTILITY VALIDATION}}\\

\multicolumn{1}{c|}{}                          & 
\multicolumn{1}{c|}{}                          & 
\textbf{VAE JS}      & \textbf{CTGAN JS}   & 
\textbf{VAE KL}        & \textbf{CTGAN KL} & 
\textbf{Real Acc}   & \textbf{VAE Synth Acc} & \textbf{CTGAN Synth Acc}\\
 
\hline
\hline
\textbf{Big data}      & 10000 & 0.115 (0.002)  & 0.136 (0.002) & 0.272 (0.014) & 0.385 (0.025) & 0.777 (0.001) & 0.748 (0.002) & 0.739 (0.006)\\
\textbf{Low data}      & 300  & 0.420 (0.003)  & 0.742 (0.003) & 1.014 (0.036) & 2.454 (0.027) & 0.775 (0.005) & 0.673 (0.006)& 0.682 (0.014) \\
\hdashline
\textbf{Pre-train}      & 300  & \textbf{0.182 (0.002)}  & \textbf{0.565 (0.002)} & \textbf{0.413 (0.008)} & \textbf{1.665 (0.025)} & N/A & \textbf{0.731 (0.005)} & 0.689 (0.003) \\
\textbf{AVG}           & 300   & \textbf{0.163 (0.003)}  & N/A           & \textbf{0.391 (0.030)} & N/A           & N/A & \textbf{0.728 (0.004)} & N/A \\
\textbf{MAML}          & 300   & \textbf{0.358 (0.004)}  & N/A           & \textbf{0.876 (0.041)} & N/A           & N/A & \textbf{0.702 (0.006)} & N/A \\
\textbf{DRS}           & 300   &\textbf{0.180 (0.002)}  & N/A           & \textbf{0.463 (0.037)} & N/A           & N/A & \textbf{0.721 (0.007)} & N/A \\
\hline
\hline
\end{tabular}}}
\caption{Adult dataset}
\vspace{0.3cm}
\end{subtable}

\begin{subtable}{\linewidth}\centering
{
\resizebox{\textwidth}{!}{%
\begin{tabular}{l|c|cccc|ccc}
\hline
\hline

\multicolumn{1}{l|}{\multirow{2}{*}{\textbf{Scenario}}} & 
\multicolumn{1}{c|}{\multirow{2}{*}{\textbf{N}}} &
\multicolumn{4}{c|}{\textbf{DIVERGENCE VALIDATION}}   & 
\multicolumn{3}{c}{\textbf{UTILITY VALIDATION}}\\

\multicolumn{1}{c|}{}                          & 
\multicolumn{1}{c|}{}                          & 
\textbf{VAE JS}      & \textbf{CTGAN JS}   & 
\textbf{VAE KL}        & \textbf{CTGAN KL} & 
\textbf{Real Acc}   & \textbf{VAE Synth MSE} & \textbf{CTGAN Synth MSE}\\
\hline
\hline
\textbf{Big data}      & 10000 & 0.742 (0.005)  & 0.746 (0.004)   & 3.243 (0.152)   & 2.920 (0.099) & 0.325 (0.005) & 0.383 (0.020)  & 0.369 (0.001) \\
\textbf{Low data}      & 300   & 0.919 (0.002)  & 0.916 (0.003)   & 5.963 (0.135)   & 6.751 (0.394) & 0.329 (0.006) & 0.516 (0.011)  & 0.756 (0.029)\\ 
 \hdashline
\textbf{Pre-train}     & 300   & \textbf{0.877 (0.002)}  & 0.948 (0.003)   & 6.215 (0.350)   & 8.394 (0.328) & N/A & \textbf{0.451 (0.012)}  & 0.767 (0.015)\\
\textbf{AVG}           & 300   & \textbf{0.724 (0.002)}  & N/A             & \textbf{3.322 (0.235)}   & N/A    & N/A   &  \textbf{0.366 (0.012)}    & N/A\\
 \textbf{MAML}          & 300   & \textbf{0.899 (0.002)}  & N/A             & 10.912 (0.472)  & N/A           & N/A    &0.686 (0.065)  & N/A\\
 \textbf{DRS}           & 300   & \textbf{0.871 (0.007)}  & N/A             & 6.118 (0.710)   & N/A        & N/A  & \textbf{0.457 (0.025)}  & N/A\\
\hline
\hline
\end{tabular}

}}

\caption{King dataset}
\vspace{0.3cm}
\end{subtable}

\begin{subtable}{\linewidth}\centering
{
\resizebox{\textwidth}{!}{%
\begin{tabular}{l|c|cccc|ccc}
\hline
\hline

\multicolumn{1}{l|}{\multirow{2}{*}{\textbf{Scenario}}} & 
\multicolumn{1}{c|}{\multirow{2}{*}{\textbf{N}}} &
\multicolumn{4}{c|}{\textbf{DIVERGENCE VALIDATION}}   & 
\multicolumn{3}{c}{\textbf{UTILITY VALIDATION}}\\

\multicolumn{1}{c|}{}                          & 
\multicolumn{1}{c|}{}                          & 
\textbf{VAE JS}      & \textbf{CTGAN JS}   & 
\textbf{VAE KL}        & \textbf{CTGAN KL} & 
\textbf{Real Acc}   & \textbf{VAE Synth Acc} & \textbf{CTGAN Synth Acc}\\
\hline
\hline
\textbf{Big data}      & 10000 & 0.395 (0.004)   & 0.635 (0.005)   & 0.911 (0.027)   & 1.678 (0.059) & 0.815 (0.007) & 0.651 (0.005)    & 0.075 (0.006)\\
\textbf{Low data}      & 300   & 0.654 (0.007)   & 0.977 (0.003)   & 1.988 (0.029)   & 10.965 (0.705) & 0.815 (0.005) & 0.410 (0.008)   & 0.039 (0.010)\\
 \hdashline
\textbf{Pre-train}     & 300   & \textbf{0.585 (0.005)}   & 0.978 (0.001)   & \textbf{1.694 (0.035)}   & \textbf{8.728 (0.650)} & N/A & 0.394 (0.010)   & 0.032 (0.006)\\
\textbf{AVG}           & 300  & \textbf{0.581 (0.007)}   & N/A             & \textbf{1.744 (0.063)}   & N/A & N/A  & 0.385 (0.005)   & N/A\\
 \textbf{MAML}          & 300   & 0.669 (0.005)   & N/A             & 2.001 (0.056)   & N/A           & N/A    &  0.318 (0.010)  & N/A\\
 \textbf{DRS}           & 300   &\textbf{0.608 (0.005) }  & N/A             & \textbf{1.635 (0.061)}   & N/A      & N/A  & \textbf{0.437 (0.012)}    & N/A\\
\hline
\hline
\end{tabular}

}}

\caption{Letter dataset}
\end{subtable}

\caption{\textbf{JS and KL results for each scenario.} `Big data' represents the ideal case with ample samples ($N=10,000$) for reliable synthetic data generation. In contrast, `Low data' simulates a more constrained setting ($N=300$). The next rows compare divergences and utility validation metrics obtained with different methodologies (pre-training, model averaging, MAML, and DRS) in the `Low data' setting. Results are shown as \textit{mean (std)}. For divergence metrics (JS, KL), lower values indicate better performance. For utility metrics, the interpretation depends on the task: for classification tasks (accuracy), higher values are better; for regression tasks (MSE), lower values are better. Bold values denote improvements due to the technique.}
\label{table:data_res}

\end{table}

\Cref{table:gains} reports the gains obtained when applying the proposed methodologies to the VAE model, measured as the absolute and relative improvement in JS and KL divergence compared to the baseline `Low data' scenario. The results show consistent improvements across most methods and datasets, with the most significant and robust gains achieved using the `AVG' strategy. While the `Pre-train' and `DRS' techniques also show positive and consistent gains in both divergence metrics, the performance of `MAML' is notably weaker and less stable. Although `MAML' yields some improvement in the Adult dataset, it fails to generalize across others and, in some cases, degrades KL divergence (e.g., King). This limited performance may stem from the nature of `MAML,' which requires a diverse set of tasks for effective meta-learning—something not fully met with only 10 training seeds per dataset. Overall, these findings confirm that the proposed inductive bias strategies enhance the fidelity of synthetic data generation, with `AVG,' `Pre-train,' and `DRS' emerging as the most effective techniques.

\renewcommand{\arraystretch}{1.25}
\begin{table}[!ht]
\centering
\resizebox{\textwidth}{!}{%
\begin{tabular}{l|cccccccc}
\hline
\hline
\multirow{2}{*}{\textbf{Dataset}} & \multicolumn{2}{c}{\textbf{Pre-train Gain}}                    & \multicolumn{2}{c}{\textbf{AVG Gain}}                          & \multicolumn{2}{c}{\textbf{MAML Gain}}                         & \multicolumn{2}{c}{\textbf{DRS Gain}}                          \\
                         & \textbf{JS}                        & \textbf{KL}                        & \textbf{JS}                        & \textbf{KL}                        & \textbf{JS}                        & \textbf{KL}                        & \textbf{JS}                        & \textbf{KL}                        \\
\hline
\hline
\textbf{Adult}                    & \textbf{0.238 (0.567) }           & \textbf{0.601 (0.593) }            & \textbf{0.257 (0.612)}           & \textbf{0.624 (0.615)}            & \textbf{0.062 (0.149)}            & \textbf{0.138 (0.136)}            & \textbf{0.240 (0.572)}         & \textbf{0.551 (0.544)}            \\

\textbf{King}                     & \textbf{0.043 (0.046)}             & -0.252 (-0.042)           & \textbf{0.195 (0.212) }         & \textbf{2.641 (0.443)}           &\textbf{0.020 (0.022) }           & -4.949 (-0.830)           & \textbf{0.048 (0.052)}          & -0.155 (-0.026)            \\

\textbf{Letter}                    & \textbf{0.069 (0.106)}            & \textbf{0.294 (0.148)  }          & \textbf{0.073 (0.112)}           & \textbf{0.243 (0.122) }           & -0.015 (-0.023)            & -0.013 (-0.007)            & \textbf{0.046 (0.070)}            & \textbf{0.352 (0.177)}            \\

\hdashline
\textbf{Average}                  & \multicolumn{1}{c}{0.117 (0.227)} & \multicolumn{1}{c}{0.214 (0.233)} & \multicolumn{1}{c}{0.175 (0.312)} & \multicolumn{1}{c}{1.169 (0.393)} & \multicolumn{1}{c}{0.023 (0.049)} & \multicolumn{1}{c}{-1.608 (-0.234)} & \multicolumn{1}{c}{0.111 (0.231)} & \multicolumn{1}{c}{0.249 (0.232)}\\
\hline
\hline\end{tabular}}
\caption{\textbf{Gains using the proposed methodology for the VAE.} Gains are represented in the following format: \textit{absolute gain (relative gain)}. The methodology achieves relative gains of up to 60\% in JS and KL divergences. Bold values indicate positive gain. Higher is better.}\label{table:gains}%
\end{table}

Interestingly, the accuracy achieved when training with a limited number of real samples closely approaches that obtained under the `Big data' configuration. This suggests that a small subset of the data may retain the most critical discriminative information required for the task. Furthermore, even in scenarios where divergence metrics indicate substantial discrepancies in the joint distribution, the utility metric (measured via accuracy) remains high. This observation implies that the generative model is effectively capturing and reproducing key informative features, which are sufficient for maintaining classification performance despite broader distributional misalignments.

Finally, it is important to acknowledge the varying computational demands of the compared methods. Model averaging is the most efficient approach, as inductive bias generation only involves calculating a mean, resulting in minimal computational overhead. In contrast, `MAML' exhibits the highest computational load due to its intricate optimization procedure. `Pre-train' and `DRS' fall between these two extremes, both requiring the training of a DGM to establish the inductive bias. Considering these findings alongside the results presented in \Cref{table:gains}, we recommend against using `MAML.' It offers minimal performance gains with significant computational costs. The other methods, on the other hand, provide a more favorable trade-off between computational efficiency and performance. Furthermore, as detailed in \ref{sec:app_exp}, reliably quantifying the benefits of our methodology in a realistic, limited-data setting is challenging. This implies that validation with a large number of samples is necessary to definitively assess which of our proposed inductive bias techniques yields superior results for a specific dataset. However, the experimental results strongly suggest potential gains that warrant further exploration.

\section{Conclusions}\label{conclusions}
This research proposed a novel approach to generate synthetic tabular data using DGMs in the context of limited datasets. Our approach leverages four distinct techniques to artificially introduce an inductive bias that guides the DGM toward generating more realistic and informative synthetic data samples. These techniques encompass two transfer learning approaches — pre-training and model averaging — and two meta-learning approaches: MAML and DRS. To facilitate the application of model averaging, MAML, and DRS, we employ the VAE model from \cite{apellaniz2024improved} and train multiple instances with different random seeds. This enables us to leverage the ensemble properties of the VAE models for techniques such as model averaging and further facilitates the application of meta-learning algorithms like MAML and DRS. We also utilized the CTGAN \citep{xu2019modeling} to evaluate pre-training and compare it with other well-known models' architectures for generating synthetic tabular data. Our approach offers several advantages over existing methods. Firstly, it effectively addresses the challenge of generating realistic synthetic data from small datasets, a common limitation in many real-world applications. Secondly, the use of transfer learning and meta-learning techniques enhances the inductive bias of the DGM, leading to more meaningful and informative synthetic data samples. However, it is also important to acknowledge the trade-offs associated with our methodology. Training VAEs with these techniques requires training multiple VAE models with different random seeds. This can lead to a significant increase in computational cost compared to simpler DGM training methods. While divergence metrics provide a valuable measure of distributional similarity, their ability to reliably assess the improvement in synthetic data quality for specific downstream tasks can be limited, especially with small datasets, as detailed in \ref{sec:app_exp}.

In conclusion, our approach offers a promising solution for generating high-quality synthetic tabular data from small datasets, particularly when VAEs apply transfer learning techniques. This work contributes to the generation of synthetic data and machine-learning applications that rely on limited data. Building on this foundation, future research should extend our methodology to other DGMs, such as diffusion models or normalizing flows, which may require adapting the bias injection techniques to new architectures. It would also be beneficial to integrate domain-specific priors or structural knowledge, especially in fields such as medicine or engineering, where expert information can enhance the fidelity of the generated results. Improving evaluation strategies beyond distributional metrics (particularly under low validation budgets) remains a key challenge, and incorporating task-aware or uncertainty-aware metrics could be a promising direction. Lastly, automating hyperparameter tuning and seed selection through meta-configuration tools or Bayesian optimization could reduce manual intervention and improve reproducibility. Overall, our work presents a general and extensible framework for enhancing synthetic data generation in low-resource tabular settings, offering a practical solution for real-world applications that require privacy-preserving and high-fidelity data synthesis.

Altogether, our methodology offers a flexible and extensible solution for generating high-fidelity synthetic data under small-sample constraints. By combining principled inductive bias strategies with scalable generative modeling, we lay the groundwork for future developments in privacy-preserving, data-efficient machine learning across domains where real data are scarce or sensitive.

\section*{Acknowledgements}
This research was supported by GenoMed4All and SYNTHEMA projects. Both have received funding from the European Union’s Horizon 2020 research and innovation program under grant agreement No 101017549 and 101095530, respectively. The authors declare that they have no known competing financial interests or personal relationships that could have appeared to influence the work reported in this paper.

\section*{Declarations}
The authors have no relevant financial or non-financial interests to disclose.

\begin{appendices}
\section{Additional Experiments under Constrained Validation Settings}\label{sec:app_exp}
To assess the impact of sample size on divergence metrics, we conducted additional validation using reduced samples for $M$ and $L$, setting $M=100$ and $L=100$ to simulate a low-data scenario. As highlighted in \citep{apellaniz2024synthetic}, sufficient samples are crucial for accurate distribution comparisons, so we expected less reliable divergence results in this setting.

\begin{table}[!ht]
\renewcommand{\arraystretch}{1.25}
\centering

\begin{subtable}{\linewidth}\centering
{
\resizebox{\textwidth}{!}{%
\begin{tabular}{l|ccc|cccc|ccc}
\hline
\hline

\multicolumn{1}{l|}{\multirow{2}{*}{\textbf{Scenario}}} & 
\multicolumn{1}{c}{\multirow{2}{*}{\textbf{N}}} &
\multicolumn{1}{c}{\multirow{2}{*}{\textbf{M}}} &
\multicolumn{1}{c|}{\multirow{2}{*}{\textbf{L}}} &
\multicolumn{4}{c|}{\textbf{DIVERGENCE VALIDATION}}   & 
\multicolumn{3}{c}{\textbf{UTILITY VALIDATION}}\\

\multicolumn{1}{c|}{}                          & 
\multicolumn{1}{c}{}                          & 
\multicolumn{1}{c}{}                          &
\multicolumn{1}{c|}{}                          &
\textbf{VAE JS}      & \textbf{CTGAN JS}   & 
\textbf{VAE KL}        & \textbf{CTGAN KL} & 
\textbf{Real Acc}   & \textbf{VAE Synth Acc} & \textbf{CTGAN Synth Acc}\\

\hline
\hline
\textbf{Big data}     & 10000 & 7500 & 1000 & 0.115 (0.002) & 0.136 (0.002) & 0.272 (0.014)  & 0.385 (0.025) &   0.777 (0.001) & 0.748 (0.002) & 0.739 (0.006) \\

\textbf{Low data}      & 300   &  100 &  100 & 0.113 (0.009) & 0.515 (0.011) & 0.773 (0.181)  & 2.694 (0.381)   &  0.726 (0.006) & 0.669 (0.004)  & 0.675 (0.007) \\
\hdashline

\textbf{Pre-train}      & 300   &  100 &  100 &\textbf{ 0.001 (0.001)} & \textbf{0.282 (0.017)} & \textbf{0.037 (0.139)}  & \textbf{1.115 (0.265)} & N/A  & \textbf{0.727 (0.003)} & \textbf{0.692 (0.004)} \\

\textbf{AVG}        & 300   &  100 &  100 &\textbf{ 0.003 (0.004)} & N/A           & \textbf{0.085 (0.076)}  & N/A       & N/A  & \textbf{0.729 (0.004)} & N/A \\

\textbf{MAML}         & 300   &  100 &  100 & 0.105 (0.004) & N/A           & \textbf{0.474 (0.099)}  & N/A    & N/A  &  \textbf{0.699 (0.005)}& N/A \\

\textbf{DRS}          & 300   &  100 &  100 & \textbf{0.001 (0.001)} & N/A           & \textbf{-0.020 (0.109)} & N/A      & N/A  & \textbf{0.726 (0.004)} & N/A \\
\hline
\hline
\end{tabular}}}

\caption{Adult dataset}
\end{subtable}

\begin{subtable}{\linewidth}\centering
{
\resizebox{\textwidth}{!}{%
\begin{tabular}{l|ccc|cccc|ccc}
\hline
\hline

\multicolumn{1}{l|}{\multirow{2}{*}{\textbf{Scenario}}} & 
\multicolumn{1}{c}{\multirow{2}{*}{\textbf{N}}} &
\multicolumn{1}{c}{\multirow{2}{*}{\textbf{M}}} &
\multicolumn{1}{c|}{\multirow{2}{*}{\textbf{L}}} &
\multicolumn{4}{c|}{\textbf{DIVERGENCE VALIDATION}}   & 
\multicolumn{3}{c}{\textbf{UTILITY VALIDATION}}\\

\multicolumn{1}{c|}{}                          & 
\multicolumn{1}{c}{}                          &
\multicolumn{1}{c}{}                          &
\multicolumn{1}{c|}{}                          &
\textbf{VAE JS}      & \textbf{CTGAN JS}   & 
\textbf{VAE KL}        & \textbf{CTGAN KL} & 
\textbf{Real MSE}   & \textbf{VAE Synth MSE} & \textbf{CTGAN Synth MSE}\\

\hline
\hline
\textbf{Big data}      & 10000 & 7500 & 1000 & 0.742 (0.005)  & 0.746 (0.004)   & 3.243 (0.152)   & 2.920 (0.099) &  0.325 (0.002) & 0.375 (0.000) & 0.369 (0.000)   \\
 \textbf{Low data}      & 300   &  100 &  100 & 0.158 (0.060)  & 0.580 (0.042)   & 1.171 (0.270)   & 4.526 (0.699) &   0.371 (0.022) & 0.520 (0.000) &  0.691 (0.000)\\
 \hdashline
  \textbf{Pre-train}      & 300   &  100 &  100 & 0.269 (0.037)  & 0.694 (0.008)   & 0.908 (0.065)   & \textbf{2.958 (0.230)} & N/A  & \textbf{0.446 (0.000)} & \textbf{0.536 (0.000)} \\
 \textbf{AVG}           & 300   &  100 &  100 & \textbf{-0.001 (0.008)} & N/A             & \textbf{0.006 (0.080) }  & N/A     & N/A  &\textbf{0.363 (0.000)} & N/A\\
 \textbf{MAML}          & 300   &  100 &  100 & 0.495 (0.051)  & N/A             & 2.516 (0.390)   & N/A   &  N/A & 0.765 (0.000) & N/A\\

 \textbf{DRS}           & 300   &  100 &  100 & 0.250 (0.032)  & N/A             & 0.944 (0.209)   & N/A    & N/A  & \textbf{0.442 (0.000)} & N/A\\
\hline
\hline
\end{tabular}}}

\caption{King dataset}
\end{subtable}

\begin{subtable}{\linewidth}\centering
{
\resizebox{\textwidth}{!}{%
\begin{tabular}{l|ccc|cccc|ccc}
\hline
\hline

\multicolumn{1}{l|}{\multirow{2}{*}{\textbf{Scenario}}} & 
\multicolumn{1}{c}{\multirow{2}{*}{\textbf{N}}} &
\multicolumn{1}{c}{\multirow{2}{*}{\textbf{M}}} &
\multicolumn{1}{c|}{\multirow{2}{*}{\textbf{L}}} &
\multicolumn{4}{c|}{\textbf{DIVERGENCE VALIDATION}}   & 
\multicolumn{3}{c}{\textbf{UTILITY VALIDATION}}\\

\multicolumn{1}{c|}{}                          & 
\multicolumn{1}{c}{}                          &
\multicolumn{1}{c}{}                          &
\multicolumn{1}{c|}{}                          &
\textbf{VAE JS}      & \textbf{CTGAN JS}   & 
\textbf{VAE KL}        & \textbf{CTGAN KL} & 
\textbf{Real Acc}   & \textbf{VAE Synth Acc} & \textbf{CTGAN Synth Acc}\\

\hline
\hline
\textbf{Big data}     & 10000 & 7500 & 1000 & 0.395 (0.004)   & 0.635 (0.005)   & 0.911 (0.027)   & 1.678 (0.059) &  0.861 (0.003) & 0.672 (0.007) & 0.066 (0.013) \\

 \textbf{Low data}      & 300   &  100 &  100 & 0.055 (0.035)   & 0.826 (0.004)   & 0.978 (0.241)   & \textbf{7.528 (1.065)} &   0.609 (0.027) & 0.401 (0.007) & 0.038 (0.004)\\
 \hdashline

  \textbf{Pre-train}      & 300   &  100 &  100 & 0.049 (0.005)   & 0.619 (0.022)   & \textbf{0.297 (0.101)}   & 4.856 (0.431)&  N/A &0.392 (0.005) & 0.040 (0.013)\\

 \textbf{AVG}           & 300   &  100 &  100 & \textbf{-0.020 (0.017)}  & N/A             & \textbf{0.322 (0.100)}   & N/A    &  N/A &0.405 (0.006) & N/A\\

 \textbf{MAML}          & 300   &  100 &  100 & 0.014 (0.045)   & N/A             & 0.845 (0.286)   & N/A    & N/A  &0.302 (0.008) & N/A\\

 \textbf{DRS}           & 300   &  100 &  100 & 0.007 (0.032)   & N/A             & \textbf{0.360 (0.090)}   & N/A     & N/A &  \textbf{0.429 (0.006) }& N/A\\
\hline
\hline
\end{tabular}}}

\caption{Letter dataset}
\end{subtable}

\caption{\textbf{Additional results for each scenario.}  
`Big data' represents the ideal case with ample samples ($N=10,000$) for reliable synthetic data generation, while `Low data' simulates a more constrained setting ($N=300$). The `Low data' rows reflect a less reliable validation case ($M=100$, $L=100$). The next rows compare metrics obtained with different methodologies (`Pre-train', `AVG', `MAML', and `DRS') in the `Low data' setting. Results are shown as \textit{mean (std)}, where lower values indicate better performance. For divergence metrics (JS, KL), lower values indicate better performance. For utility metrics, the interpretation depends on the task: for classification tasks (accuracy), higher values are better; for regression tasks (MSE), lower values are better. Bold values denote improvements due to the technique. A slightly negative divergence value close to zero is acceptable, as it reflects an estimation error from the divergence approximation.}\label{table:constrained_results}%
\end{table}

\Cref{table:constrained_results} shows that, under constrained validation conditions ($M=100$, $L=100$), the improvements observed with our proposed techniques are less consistent across datasets. Divergence values, particularly for JS, remain artificially low, suggesting that small validation sets lack the statistical power to capture meaningful distributional differences. This can lead to misleading conclusions about the similarity between real and synthetic distributions. In contrast, utility validation remains relatively stable across most cases, showing similar trends to those observed with larger $M$ and $L$ values. This indicates that the generative models are able to capture the key informative features needed for classification or regression, even when the global distributional alignment cannot be reliably assessed due to sample scarcity.

\section{Additional Datasets}\label{app:additional_data}
In addition to the three main datasets described in \ref{sec:data}, we also evaluated our methodology on two supplementary public datasets. These additional datasets provide further evidence of the generalizability of our approach across diverse domains and data modalities. Their inclusion reinforces the robustness of our findings beyond the core evaluation set. Below, we describe each of them:

\begin{itemize}
    \item \textbf{News:} The News Popularity Prediction dataset \citep{news} includes 39,644 samples and 58 features from articles published on the Mashable news blog. It is a multivariate dataset with continuous and categorical variables that predict article popularity based on social media shares.
     
     \item \textbf{Intrusion:} The KDD Cup 1999 Data \citep{intrusion} consists of 494,021 samples and 39 features for classifying connections in a military network environment. It contains categorical, integer, and binary attributes and was used in The Third Knowledge Discovery and Data Mining Competition.
   
\end{itemize}

These datasets were selected to diversify the experimental setting in terms of sample size, number of features, and classification difficulty. As with the primary datasets, they were used to test both divergence-based and utility-based evaluations under different data scarcity scenarios. Results for these datasets can be found in \Cref{table:add_datasets_results}.

\begin{table}[!ht]
\renewcommand{\arraystretch}{1.25}
\centering

\begin{subtable}{\linewidth}\centering
{
\resizebox{\textwidth}{!}{%
\begin{tabular}{l|ccc|cccc|ccc}
\hline
\hline

\multicolumn{1}{l|}{\multirow{2}{*}{\textbf{Scenario}}} & 
\multicolumn{1}{c}{\multirow{2}{*}{\textbf{N}}} &
\multicolumn{1}{c}{\multirow{2}{*}{\textbf{M}}} &
\multicolumn{1}{c|}{\multirow{2}{*}{\textbf{L}}} &
\multicolumn{4}{c|}{\textbf{DIVERGENCE VALIDATION}}   & 
\multicolumn{3}{c}{\textbf{UTILITY VALIDATION}}\\

\multicolumn{1}{c|}{}                          & 
\multicolumn{1}{c}{}                          & 
\multicolumn{1}{c}{}                          &
\multicolumn{1}{c|}{}                          &
\textbf{VAE JS}      & \textbf{CTGAN JS}   & 
\textbf{VAE KL}        & \textbf{CTGAN KL} & 
\textbf{Real Acc}   & \textbf{VAE Synth Acc} & \textbf{CTGAN Synth Acc}\\

\hline
\hline
\textbf{Big data}      & 10000 & 7500 & 1000 & 0.223 (0.002)   & 0.484 (0.010) & 0.525 (0.025)   & 1.432 (0.066) & 0.699 (0.007) & 0.622 (0.002) & 0.616 (0.008) \\
 \textbf{Low data}      & 300 & 7500 & 1000 & 0.831 (0.004)   & 0.966 (0.002) & 4.301 (0.114)   & 10.961 (0.317) & 0.696 (0.009) & 0.564 (0.016) &  0.470 (0.030)\\
 \textbf{Low data}      & 300   &  100 &  100 & 0.182 (0.022)   & 0.864 (0.017) & 0.460 (0.120)   & 5.673 (0.546)  & 0.555 (0.043) & 0.576 (0.009) &  0.512 (0.028)\\
 \hdashline
 
 \textbf{Pre-train}      & 300   & 7500 & 1000 & \textbf{0.734 (0.005)}   & \textbf{0.935 (0.001)} & \textbf{3.403 (0.098) }  & 10.441 (0.544)& N/A & \textbf{0.596 (0.003)} & \textbf{0.525 (0.015)} \\
 \textbf{Pre-train}      & 300   &  100 &  100 & \textbf{0.009 (0.009)}   & \textbf{0.701 (0.010)} & \textbf{0.060 (0.088) }  & \textbf{3.868 (0.570) } & N/A&  \textbf{0.595 (0.004)} & 0.513 (0.023)  \\
 
  \textbf{AVG}           & 300   & 7500 & 1000 & \textbf{0.618 (0.003)}   & N/A           & \textbf{2.542 (0.054) }  & N/A         & N/A& \textbf{0.588 (0.002)} & N/A \\
 \textbf{AVG}           & 300   &  100 &  100 & \textbf{-0.001 (0.004)}  & N/A           & \textbf{-0.101 (0.114) } & N/A   & N/A&  \textbf{0.587 (0.002)}& N/A \\
 
 \textbf{MAML}          & 300   & 7500 & 1000 & 0.831 (0.008)   & N/A           & 4.381 (0.129)   & N/A         & N/A & 0.577 (0.008)  & N/A \\
 \textbf{MAML}          & 300   &  100 &  100 & \textbf{0.002 (0.003)}   & N/A           & \textbf{0.101 (0.132)}   & N/A      & N/A & 0.580 (0.006)& N/A \\
 
 \textbf{DRS}           & 300   & 7500 & 1000 & \textbf{0.603 (0.015)}   & N/A           & \textbf{2.467 (0.050)}   & N/A               & N/A & \textbf{0.605 (0.011)} &  N/A\\
 \textbf{DRS}            & 300   &  100 &  100 & \textbf{0.026 (0.006)}   & N/A           & \textbf{0.034 (0.120)}   & N/A        & N/A & \textbf{0.609 (0.010)}&  N/A\\
 
\hline
\hline
\end{tabular}}}

\caption{News dataset}
\vspace{0.3cm}
\end{subtable}

\begin{subtable}{\linewidth}\centering
{
\resizebox{\textwidth}{!}{%
\begin{tabular}{l|ccc|cccc|ccc}
\hline
\hline

\multicolumn{1}{l|}{\multirow{2}{*}{\textbf{Scenario}}} & 
\multicolumn{1}{c}{\multirow{2}{*}{\textbf{N}}} &
\multicolumn{1}{c}{\multirow{2}{*}{\textbf{M}}} &
\multicolumn{1}{c|}{\multirow{2}{*}{\textbf{L}}} &
\multicolumn{4}{c|}{\textbf{DIVERGENCE VALIDATION}}   & 
\multicolumn{3}{c}{\textbf{UTILITY VALIDATION}}\\

\multicolumn{1}{c|}{}                          & 
\multicolumn{1}{c}{}                          & 
\multicolumn{1}{c}{}                          &
\multicolumn{1}{c|}{}                          &
\textbf{VAE JS}      & \textbf{CTGAN JS}   & 
\textbf{VAE KL}        & \textbf{CTGAN KL} & 
\textbf{Real Acc}   & \textbf{VAE Synth Acc} & \textbf{CTGAN Synth Acc}\\

\hline
\hline

\textbf{Big data}      & 10000 & 7500 & 1000 & 0.795 (0.014)   & 0.459 (0.002) & 3.094 (0.113) & 1.724 (0.106) & 0.742 (0.369) & 0.330 (0.178) & 0.590 (0.096)\\
\textbf{Low data}      & 300   & 7500 & 1000 & 0.936 (0.003)   & 0.969 (0.004) & 8.297 (0.516) & 9.877 (0.770) & 0.844 (0.173) & 0.567 (0.107) &  0.733 (0.008)\\
\textbf{Low data}      & 300   &  100 &  100 & 0.228 (0.033)   & 0.506 (0.021) & 1.600 (1.133) & 4.353 (0.610) &  0.917 (0.010) & 0.575 (0.036)  & 0.735 (0.025)  \\

 \hdashline
\textbf{Pre-train}     & 300   & 7500 & 1000 & \textbf{0.797 (0.008)}   & 0.965 (0.001) & \textbf{4.103 (0.243)} & 8.409 (0.842) & N/A & \textbf{0.756 (0.022)} & 0.755 (0.016)\\
\textbf{Pre-train}       & 300   &  100 &  100 & \textbf{0.066 (0.014)}   & 0.502 (0.044) & \textbf{0.236 (0.076)} & 3.680 (0.333) &  N/A &  \textbf{0.780 (0.011)} &  0.759 (0.013) \\

\textbf{AVG}           & 300   & 7500 & 1000 & \textbf{0.850 (0.002)}   & N/A           & \textbf{4.934 (0.134)} & N/A   & N/A & \textbf{0.801 (0.007)} &   N/A  \\
\textbf{AVG}      & 300   &  100 &  100 & \textbf{0.058 (0.003)}   & N/A           & \textbf{0.330 (0.114)} & N/A  &  N/A &  \textbf{0.798 (0.006)} &  N/A\\

 \textbf{MAML}          & 300   & 7500 & 1000 & 0.945 (0.005)   & N/A           & 8.046 (0.457) & N/A      & N/A & 0.652 (0.061) &  N/A\\
\textbf{MAML}      & 300   &  100 &  100 & 0.368 (0.023)   & N/A           & 3.131 (0.555) & N/A   & N/A  & 0.532 (0.145) &  N/A \\
 
 \textbf{DRS}           & 300   & 7500 & 1000 & \textbf{0.844 (0.015)}   & N/A           & \textbf{5.613 (0.435)} & N/A     & N/A & 0.544 (0.242) & N/A\\
\textbf{DRS}       & 300   &  100 &  100 & \textbf{0.043 (0.011)}   & N/A           & \textbf{0.289 (0.187)} & N/A   & N/A  & 0.585 (0.121) &   N/A\\

\hline
\hline
\end{tabular}}}
\caption{Intrusion dataset}

\end{subtable}

\caption{\textbf{Results for additional datasets in each scenario.} `Big data' represents the ideal case with ample samples ($N=10,000$) for reliable synthetic data generation, while `Low data' simulates a more constrained setting ($N=300$). Two `Low data' rows reflect different validation sample sizes: a more reliable case ($M=7,500$, $L=1,000$) and a less reliable case ($M=100$, $L=100$), which apply to the following rows. The next rows compare metrics obtained with different methodologies (`Pre-train,' `AVG,' `MAML,' and `DRS') in the `Low data' setting. Results are shown as \textit{mean (std)}, where lower values indicate better performance. For divergence metrics (JS, KL), lower values indicate better performance. For utility metrics, the interpretation depends on the task: for classification tasks (accuracy), higher values are better; for regression tasks (MSE), lower values are better. Bold values denote improvements due to the technique. A slightly negative divergence value close to zero is acceptable, as it reflects an estimation error from the divergence approximation.}

\label{table:add_datasets_results}%

\end{table}

\Cref{table:add_datasets_results} confirms the generalizability of our findings across the additional News and Intrusion datasets. Consistent with the main experimental results, the VAE model outperforms CTGAN in both divergence and utility metrics across most scenarios. Furthermore, the proposed techniques—particularly `Pre-train,' `AVG,' and `DRS'—consistently improve over the baseline (`Low data') in terms of JS and KL divergence, as well as downstream performance (accuracy). The only exception remains `MAML,' which shows less stable behavior and, in several cases, fails to provide improvements. These results reinforce the robustness of our methodology and its applicability to diverse tabular domains.

\section{Computational Resources and Training Time Analysis}\label{app:times}
All experiments were conducted on a prosumer-grade workstation equipped with an AMD Ryzen Threadripper PRO 5975WX processor featuring 32 cores and 64 threads. This CPU operates at a base frequency of 3.6 GHz with a boost up to 4.5 GHz, offering substantial parallel computing capacity. The machine was configured with 128 GB of DDR4 RAM and a 1 TB NVMe SSD for fast data access, along with additional storage on dual 4 TB HDDs. Although the workstation included an NVIDIA RTX 4090 graphics card, GPU acceleration was not required for training the VAE-based model, as all experiments involving the VAE were performed on the CPU. On the other hand, the CTGAN implementation used in this work utilizes GPU acceleration by default.

Despite the added complexity introduced by some of the proposed techniques, training times remained practical for both models. \Cref{table:runtime} summarizes the average training durations for each dataset and methodology over 10 runs using different random seeds. The VAE, although trained with 10 different seeds, achieved competitive training times. Even in the most demanding scenario (`Big data'), the average training time across datasets remained below 37 minutes (2,198 seconds), and no single experiment exceeded one hour. CTGAN experiments were significantly faster, especially due to GPU usage, yet both architectures can be considered efficient and lightweight in practice. Importantly, for both models, applying the proposed methodologies in the low-data regime, such as `Pre-train,' `AVG,' `MAML,' and `DRS,' results in only moderate increases in training time compared to the baseline `Low data' scenario. This demonstrates the practicality and scalability of our approach: the benefits of inductive bias integration are achieved with minimal additional computational cost, making the methods suitable for real-world settings where time and resources are constrained.

\begin{table}[!ht]
\renewcommand{\arraystretch}{1.15}
\centering
\resizebox{\textwidth}{!}{
\begin{tabular}{l|cccccccccccc}
\hline\hline
\multirow{2}{*}{\textbf{Dataset}} & \multicolumn{2}{c}{\textbf{Big data}}                    & \multicolumn{2}{c}{\textbf{Low data}}                          & \multicolumn{2}{c}{\textbf{Pre-train}}                         & \multicolumn{2}{c}{\textbf{AVG}}        & \multicolumn{2}{c}{\textbf{MAML}}    & \multicolumn{2}{c}{\textbf{DRS}}              \\
                         & \textbf{VAE}                        & \textbf{CTGAN}                        & \textbf{VAE}                        & \textbf{CTGAN}                        & \textbf{VAE}                        & \textbf{CTGAN}                       & \textbf{VAE}                        & \textbf{CTGAN} & \textbf{VAE}                        & \textbf{CTGAN}  & \textbf{VAE}                        & \textbf{CTGAN}                    \\

\hline\hline
\textbf{Adult}     & 1893.25  &   382.356 &        61.6617 &          19.0268 &       1367.84  &          402.089 &  145.509  & ---         &    125.314 & ---          &  463.661  & ---         \\
\textbf{News}       & 7248.41  &   532.651 &       115.353  &          27.1911 &       4593.78  &          558.03  &  139.851  & ---         &    534.483 & ---          &  386.022  & ---         \\
\textbf{King}      & 1282.78  &   393.782 &        70.4384 &          20.6333 &        567.242 &          416.421 &   75.209  & ---         &    204.017 & ---          &  133.523  & ---         \\
\textbf{Intrusion} & 1651.26  &   570.559 &        82.2507 &          29.3055 &       4909.34  &          604.459 &  187.802  & ---         &    234.104 & ---          &  923.386  & ---         \\
\textbf{Letter}    &  821.65  &   299.236 &        96.9268 &          16.5833 &        776.11  &          323.648 &   84.3122 & ---         &    193.772 & ---          &  210.785  & ---         \\
\hline\hline
\textbf{Average}   & 2579.87  &   435.317 &        85.3269 &          22.94798 &       2442.06  &          460.929 &  126.537  & ---         &    258.338 & ---          &  423.875  & ---         \\
\hline\hline
\end{tabular}
}
\caption{Average execution times (in seconds) across 10 runs with different random seeds for the various scenarios and datasets.}
\label{table:runtime}
\end{table}

Overall, this analysis demonstrates that the benefits provided by inductive bias techniques come at a relatively low computational cost. Even when training multiple VAE instances or applying more complex optimization procedures, the observed increases in runtime remain within practical limits. This reinforces the applicability of our methods in real-world, resource-constrained environments and supports their use in privacy-sensitive domains where synthetic data are needed, but GPU availability may be limited.

\end{appendices}

\bibliographystyle{unsrt}  
\bibliography{main}

\end{document}